\documentclass[a4paper,fleqn]{cas-dc}

\usepackage[numbers]{natbib}
\usepackage{eurosym,makecell,comment}
\usepackage{multirow}
\usepackage{threeparttable}
\usepackage{framed}
\usepackage{hyperref}
\usepackage{url}
\usepackage{verbatim}
\usepackage{graphicx}
\usepackage{subcaption}
\usepackage{float}
\usepackage{xcolor}
\usepackage{microtype}
\usepackage{colortbl}
\usepackage{amsmath,amssymb,amsfonts}
\usepackage{tikz}
\usepackage{pgfplots}
\usepackage{listings}
\usepackage{array}
\usepackage{ragged2e}

\usepackage{pifont}
\usepackage{algorithm}
\usepackage{algpseudocode}
\usepackage{balance}

\usepackage{soul}

\usepackage[german,english]{babel}
\usepackage[utf8]{inputenc}
\usepackage{latexsym}
\usepackage{epsfig}
\usepackage{moreverb}
\usepackage{rotating}
\usepackage{enumerate}
\usepackage{graphics,wrapfig}
\usepackage{url}
\usepackage{forest}
\usepackage{listings, lstautogobble}
\usepackage{xspace}
\usepackage{multirow}

\usepackage{multicol}
\usepackage{array}
\usepackage{tabularx} 
\usepackage{algorithm}
\usepackage{algpseudocode}
\usepackage{subcaption}
\usepackage{makecell}
\usepackage{hhline}
\usepackage{lipsum}
\definecolor{Gray}{gray}{0.95}
\newcolumntype{g}{>{\columncolor{Gray}}p}

\usepackage{booktabs}

\usetikzlibrary{calc}

\pgfplotsset{compat=1.18} 

\def \1{\textit{(i)}}
\def \2{\textit{(ii)}}
\def \3{\textit{(iii)}}
\def \4{\textit{(iv)}}
\def \5{\textit{(v)}}

\usepackage{etoolbox}

\setstcolor{blue}

\newtoggle{removeItalic}
\togglefalse{removeItalic} 
\iftoggle{removeItalic} {
    \renewcommand{\textit}[1]{#1}
}

\begin{document}
\let\WriteBookmarks\relax
\def\floatpagepagefraction{1}
\def\textpagefraction{.001}
\shorttitle{Automotive HIL Test Case Automation}
\shortauthors{Feng et~al.}

\title[mode = title]{Smarter, not Bigger: Fine-Tuned RAG-Enhanced LLMs for Automotive HIL Testing}

\author[1]{Chao {Feng}*}[orcid=0000-0002-0672-1090]
\author[1]{Zihan {Liu}}[]
\author[2]{Siddhant {Gupta}}[]
\author[2]{Gongpei {Cui}}[]
\author[1]{Jan {von der Assen}}[orcid=0000-0002-0591-8887]
\author[1]{Burkhard {Stiller}}[orcid=0000-0002-7461-7463]

\address[1]{Communication Systems Group CSG, Department of Informatics IfI, University of Zurich UZH, 8050 Zürich, Switzerland}
\address[2]{Volvo Car Corporation, 405 31 Göteborg, Sweden}

\cortext[cor1]{Corresponding author.
Email address: cfeng@ifi.uzh.ch (C. Feng), zihan.liu@uzh.ch (Z. Liu), siddhant.gupta@volvocars.com (S. Gupta), gongpei.cui@volvocars.com (G. Cui), vonderassen@ifi.uzh.ch (J. von der Assen), stiller@ifi.uzh.ch (B. Stiller)}


\maketitle

\begin{abstract}
Hardware-in-the-Loop (HIL) testing is essential for automotive validation but suffers from fragmented and underutilized test artifacts. This paper presents HIL-GPT, a retrieval-augmented generation (RAG) system integrating domain-adapted large language models (LLMs) with semantic retrieval. HIL-GPT leverages embedding fine-tuning using a domain-specific dataset constructed via heuristic mining and LLM-assisted synthesis, combined with vector indexing for scalable, traceable test case and requirement retrieval. Experiments show that fine-tuned compact models, such as \texttt{bge-base-en-v1.5}, achieve a superior trade-off between accuracy, latency, and cost compared to larger models, challenging the notion that bigger is always better. An A/B user study further confirms that RAG-enhanced assistants improve perceived helpfulness, truthfulness, and satisfaction over general-purpose LLMs. These findings provide insights for deploying efficient, domain-aligned LLM-based assistants in industrial HIL environments.
\end{abstract}

\section{Introduction}
Hardware and software testing is a critical phase in the automotive production lifecycle, ensuring that components function as specified and that the vehicle meets reliability and safety standards~\cite{garikapati2024dual}. As systems grow more complex, manual testing becomes increasingly impractical due to high costs, time demands, and safety risks, particularly when evaluating failure modes or edge cases~\cite{gaspar2024review}. To address these challenges, the industry has adopted Hardware-in-the-Loop (HIL) testing, which emulates real-world operating conditions by supplying simulated inputs to the system under test. HIL improves efficiency and cost-effectiveness while enabling safe evaluation of critical scenarios under controlled conditions~\cite{cheng2024survey}.

Building on these advantages, HIL has become a cornerstone of contemporary automotive development, supporting the validation of increasingly sophisticated functionalities across perception, decision-making, and actuation layers~\cite{howick2024framework}. By integrating sensor emulation (e.g., LiDAR, radar, and camera) and dynamic traffic scenario engines, modern HIL configurations provide realistic, interactive environments that enable thorough testing of safety-critical systems under diverse conditions~\cite{jooriah2024co}. 

While HIL enables realistic and efficient validation, its practical implementation relies on a complex ecosystem of communication protocols, scripting languages, and testing platforms~\cite{ali2024efficient}. At the heart of this ecosystem, the Controller Area Network (CAN) protocol serves as the communication backbone of embedded automotive systems, facilitating message exchange between electronic control units (ECUs)~\cite{inproceedings}. Complementing this, the Communication Access Programming Language (CAPL) has become the standard scripting language for defining event-driven interactions, timing constraints, and fault injection scenarios within HIL environments~\cite{annilsson_can}. These tools are typically integrated into platforms like CANoe, supporting both virtual and mixed physical-virtual test configurations.

Despite the increasing maturity of HIL tools and workflows, significant challenges remain in practice. Automotive validation environments still rely on tightly coupled and heterogeneous infrastructure, where critical metadata is scattered across non-standard formats and disconnected systems~\cite{cheng2024survey}. Over time, manufacturers have accumulated vast repositories of test cases and validation artifacts, yet much of this data remains underutilized due to its fragmented and unstructured nature. This underutilization of valuable engineering knowledge highlights the need for intelligent methods that can extract, organize, and leverage legacy test data to enhance the efficiency, scalability, and consistency of HIL-based validation processes.

Recent advances in Artificial Intelligence (AI), particularly intelligent and expert systems powered by Large Language Models (LLMs), have demonstrated potential in supporting software and hardware testing~\cite{jin2024jailbreakzoo}. Although LLM-based intelligent systems have been widely applied in knowledge retrieval and automated test generation tasks, their integration into HIL validation remains underexplored. The proprietary, closed nature of HIL platforms and the highly specialized knowledge they require pose significant challenges for general-purpose LLMs~\cite{ji2023survey}. Addressing these domain-specific limitations and unlocking the potential of accumulated test data for HIL workflows thus presents a critical and largely unaddressed research problem, which this work tackles.

\begin{table*}[ht]
\centering
\caption{Summary of Related Work on LLMs in Domain-Specific Engineering}
\label{tab:fine-tuning-rag}
\footnotesize
\begin{tabularx}{\textwidth}{l l l X X}
\toprule
\textbf{Work} & \textbf{Year} & \textbf{Domain} & \textbf{Methodology} & \textbf{Key Findings} \\
\midrule
\multicolumn{4}{l}{\textit{\textbf{Fine-Tuning}}} \\
\hline
\cite{elhambakhsh2025domain} & 2025 & Mechanical design annotation & Fine-tuned GPT-3.5 on functional component classification data & Improved annotation accuracy through domain adaptation \\
\cite{lu2025fine}  & 2025 & Multi-domain adaptation & Compared CPT, SFT, DPO fine-tuning strategies & Multi-strategy fine-tuning yields synergistic benefits \\
\cite{englhardt2024exploring}  & 2024 & Embedded coding & Evaluated general-purpose LLMs on hardware interface coding & Strong reasoning, but poor reliability in low-level code \\
\midrule
\multicolumn{4}{l}{\textit{\textbf{RAG}}} \\
\hline
\cite{ZHANG2023110264}  & 2023 & Industrial QA & Dual-encoder RAG: dense retriever + GPT-style generator trained on QA corpus & Improved QA accuracy and interpretability \\
\cite{hu2024rag}  & 2024 & Aerospace diagnostics & RAG-based transformer with fast keyword retriever + ranker grounded in fault descriptions & Improved factuality, reduced hallucinations \\
\cite{yu2025finemedlm}  & 2025 & Embedded diagnostics & Domain-pretrained RAG with weakly labeled triplet-tuned retriever & Addressed terminology misalignment; retrieval latency remains a bottleneck \\
\bottomrule
\end{tabularx}
\end{table*}

To address the domain-specific challenges of applying general-purpose LLMs in automotive HIL testing, this work proposes HIL-GPT, an intelligent agent that leverages fine-tuning and retrieval-augmented generation (RAG) to deliver contextualized access to requirements, test sequences, and CAN signal metadata. Central to this approach is a hybrid question–answer (QA) dataset, constructed from both public and proprietary sources and enriched with positive and negative QA pairs to improve intent recognition and domain understanding. The main contributions of this work are summarized as follows:
\begin{itemize}
    \item A modular architecture that integrates fine-tuned LLMs with a vector-based retrieval backend to enable contextualized querying of automotive testing knowledge through structured interfaces.
    
    \item A hybrid, domain-specific dataset combining curated public examples with internal automotive specifications, CAPL scripts, and CAN definitions, augmented with synthetic positive and negative QA pairs to improve intent understanding.
    
    \item A domain-adapted agent, named HIL-GPT, integrated into the HIL testing process and designed to interact with existing tools and workflows, supported by strategies for dataset maintenance and sustainable integration. The proposed HIL-GPT agent was deployed in a real-world testing environment and evaluated by professional engineers in collaboration with an automotive manufacturer.
    
    \item Comprehensive evaluations, combining empirical measurements of embedding quality, hallucination mitigation, and model scaling effects within the RAG pipeline under domain-specific constraints, were conducted alongside qualitative A/B testing with domain experts in a real automotive HIL environment. This expert-driven evaluation not only validated the system’s practical utility in industrial workflows but also yielded actionable insights on retrieval–generation alignment, response scoping, and usability priorities for field deployment.
\end{itemize}

The remainder of this paper is structured as follows. Section~\ref{sec:related}  contains findings from the literature review on AI-assisted test generation and domain-specific LLMs. Section~\ref{sec:architecture} describes the architecture of the proposed solution, outlining key components and their interactions.  Section~\ref{sec:evaluation} presents the experimental evaluation on real HIL configurations. Section~\ref{sec:dis} discusses the key insights found by this work. Finally, Section~\ref{sec:summary} concludes the paper and discusses directions for future work.

\section{Background and Related Work}
\label{sec:related}
This section outlines the technical background and situates the present work within the existing literature. It first briefly introduces HIL testing and the CAN protocol as foundational components of automotive validation workflows. It then reviews recent developments in the application of LLMs to domain-specific tasks, highlighting the lack of prior research on their integration into HIL environments.

\subsection{HIL and CAN in Automotive Testing}
HIL testing is a widely adopted methodology in the automotive industry, enabling real hardware components to interact with simulated environments for validation purposes. HIL test benches are indispensable for verifying controllers and sensors under controlled, repeatable conditions before deployment. They simulate diverse operating scenarios, including steering inputs, wheel speeds, display operations, and sensor states, providing comprehensive coverage at lower cost and risk compared to on-road tests~\cite{garikapati2024dual, cheng2024survey}. HIL systems also support fault injection, regression, and boundary testing, improving reliability and test coverage~\cite{gaspar2024review}.

A typical HIL configuration includes a real-time simulation platform, the device under test (DUT), signal conditioning hardware, and orchestration software~\cite{howick2024framework}. These components operate deterministically and maintain closed-loop feedback, which is critical for meeting real-time constraints. HIL setups are often complemented by sensor emulation (LiDAR, radar, cameras) and traffic scenario engines to test safety-critical components~\cite{jooriah2024co}. Furthermore, industry-standard frameworks like AUTOSAR are routinely validated through HIL platforms to ensure vendor compliance~\cite{kim2024autosar}.

The CAN, developed by Bosch in the 1980s, serves as the communication backbone for automotive embedded systems. It facilitates message exchange between electronic control units (ECUs) via standardized message IDs and OEM-specific encoding rules~\cite{bosch_can20}. Tools such as CANoe extend CAN capabilities, enabling fully virtual or hybrid simulations and supporting the full lifecycle of network design, simulation, and testing~\cite{ali2024efficient, inproceedings}. CAPL scripting is commonly used alongside CAN to simulate, monitor, and inject events in test environments, enabling reproducible, event-driven validation workflows~\cite{vector_capl}.

However, these workflows are highly heterogeneous and fragmented. Metadata, such as test objectives, signal mappings, and feature tags, is stored in inconsistent formats across disconnected tools, complicating automation and reuse. The tight coupling between CAN, CAPL, and proprietary infrastructure presents barriers to integrating modern AI techniques into HIL processes.

\subsection{LLMs in Domain-Specific Engineering}
As no prior research has explored the use of LLMs in HIL testing, this section reviews how LLMs have been adapted to domain-specific engineering contexts to draw insights for our approach. Recent advances have applied two main strategies: fine-tuning and RAG. Fine-tuning adapts a model’s internal weights using curated corpora to internalize domain-specific terminology and logic~\cite{elhambakhsh2025domain, lu2025fine}, while RAG combines a frozen or lightly tuned model with an external retriever that provides relevant context at inference time~\cite{ZHANG2023110264, hu2024rag}. As shown in \tablename~\ref{tab:fine-tuning-rag}, both strategies have demonstrated promising results in technical and industrial settings.

\begin{table*}[h]
\centering
\caption{Related Work on Data Construction and Embedding Optimization for Domain-Specific LLMs}
\label{tab:data-embedding}
\footnotesize
\begin{tabularx}{\textwidth}{l l l X X}
\toprule
\textbf{Work} & \textbf{Year} & \textbf{Domain / Task} & \textbf{Methodology} & \textbf{Key Findings} \\
\midrule
\cite{gao2021simcse}  & 2021 & Sentence embedding & Triplet-based contrastive learning (SimCSE) & High-quality embeddings without task-specific labels \\
\cite{yang2024rethinking}  & 2024 & Systems engineering & Engineer-in-the-loop triplet construction & Higher-quality triplets for safety-critical settings \\
\cite{liu2024optimizing}  & 2024 & Automotive RAG & Layout-aware chunking and metadata alignment & Improved retrieval precision \\
\cite{nguyen2024enhancing}  & 2024 & RAG embedding fine-tuning & Studied embedding vs. LLM fine-tuning & Embedding fine-tuning more impactful \\
\cite{zhou2025representation}  & 2025 & Representation learning & Hierarchical embedding fine-tuning & Better adaptation with limited data \\
\cite{zhang2025hallucination}  & 2025 & Hallucination mitigation & Model-augmented embedding fine-tuning & Reduced hallucinations in retrieval-heavy tasks \\
\bottomrule
\end{tabularx}
\end{table*}

Fine-tuning pipelines have been widely explored to improve domain adaptation. \cite{elhambakhsh2025domain} reported that fine-tuning GPT-3.5 on mechanical design annotations led to higher functional component classification accuracy. Building on this line of work, \cite{lu2025fine} compared continued pretraining, supervised fine-tuning, and direct preference optimization, and observed that hybrid strategies can produce complementary gains. In the embedded development domain, \cite{englhardt2024exploring} noted that LLMs often misinterpret hardware protocols, highlighting the importance of domain-targeted fine-tuning.

In parallel, RAG-based approaches have gained traction in engineering applications. One example is the dual-encoder framework proposed by \cite{ZHANG2023110264}, which retrieves specification and code snippets to pair with user queries, improving both accuracy and interpretability. Similarly, \cite{hu2024rag} applied a keyword-driven retriever with a transformer ranker to aerospace diagnostics, grounding responses in fault descriptions to mitigate hallucinations. Extending this idea to embedded microcontroller diagnostics, \cite{yu2025finemedlm} fine-tuned a domain-pretrained retriever on weakly labeled triplets, achieving better terminology alignment, though at the cost of increased retrieval latency.

A recurring challenge across these studies is the scarcity and heterogeneity of labeled domain data. To mitigate this, weak supervision and contrastive learning have emerged as practical solutions (\tablename~\ref{tab:data-embedding}). Early work such as SimCSE~\cite{gao2021simcse} demonstrated that high-quality sentence embeddings can be learned without task-specific labels through triplet-based contrastive objectives. Building on this foundation, \cite{yang2024rethinking} incorporated engineer-in-the-loop heuristics to generate more informative triplets in safety-critical environments. Complementary to these sampling strategies, \cite{liu2024optimizing} and \cite{nguyen2024enhancing} highlighted that enhancing embedding quality, via techniques such as layout-aware chunking and embedding fine-tuning, can yield larger gains in retrieval accuracy than direct LLM fine-tuning. More recently, advanced adaptation methods, including hierarchical embedding refinement~\cite{zhou2025representation} and model-augmented representation learning~\cite{zhang2025hallucination}, have shown particular promise in low-resource settings by improving retrieval relevance and reducing hallucinations. Together, these approaches underscore that in domain-specific RAG systems, targeted embedding optimization often offers a better return under data-scarce constraints.

While these approaches illustrate the feasibility of adapting LLMs to technical domains, none address the unique requirements of HIL workflows. Specifically, existing work does not consider the heterogeneity of HIL data sources and the integration of CAN and CAPL artifacts. This work fills these gaps by proposing a domain-adapted agent that combines fine-tuned embeddings and an RAG pipeline, tailored to the specialized data structures and operational constraints of HIL testing, while attesting to its efficacy by testing in a real setting.

\section{Architecture of HIL-GPT}
\label{sec:architecture}
\begin{figure*}
    \centering
    \includegraphics[width=\linewidth]{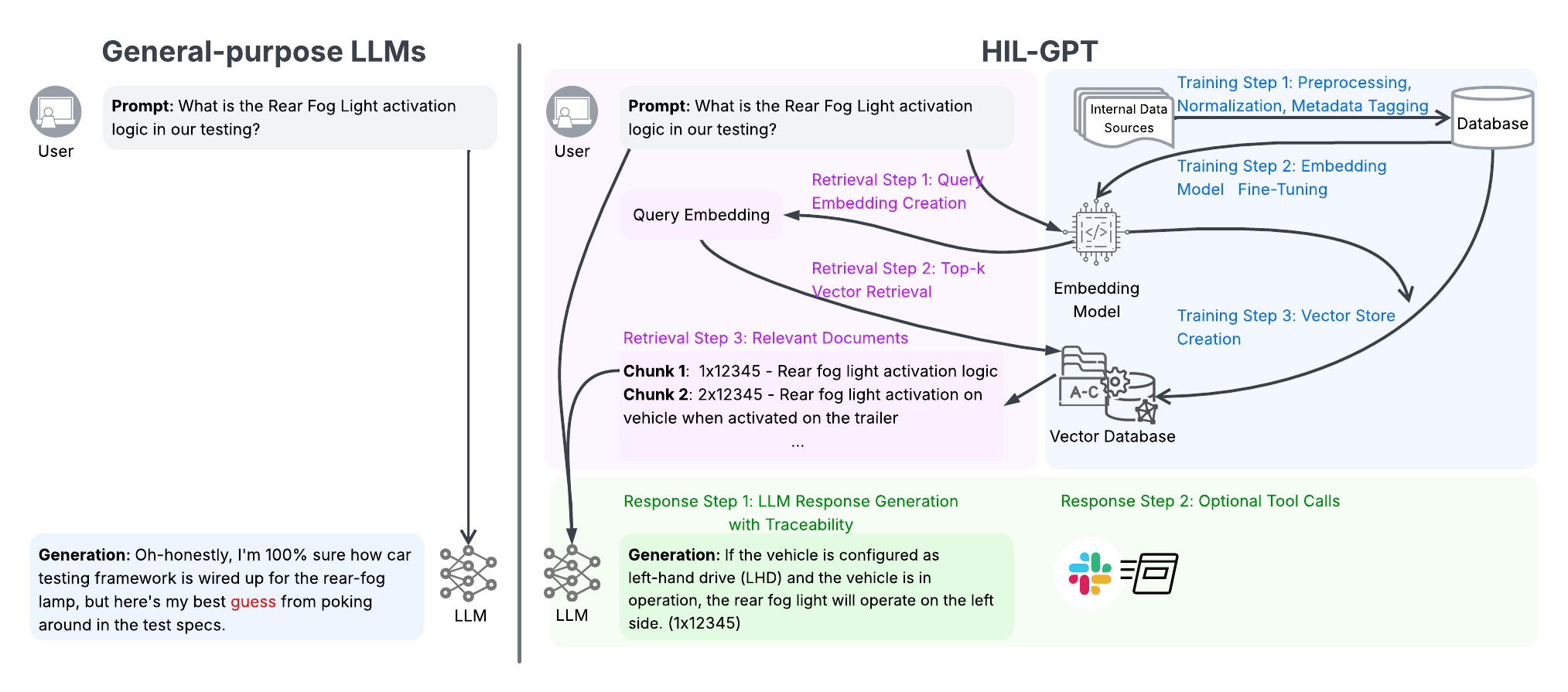}
    \caption{System architecture of HIL-GPT.}
    \label{fig:arch}
\end{figure*}

This section presents the detailed architecture of HIL-GPT, an intelligent assistant system tailored for supporting requirement interpretation and test sequence retrieval in automotive HIL environments. The architecture addresses the integration of domain-specific knowledge, fragmented data management, and the stringent operational constraints of real-world industrial settings. It consists of two distinct but interconnected pipelines: the offline knowledge curation pipeline and the online inference and response generation pipeline.  \figurename~\ref{fig:arch} illustrates the overall architecture and contrasts it with a baseline general-purpose LLM.

\subsection{Offline Pipeline: Knowledge Curation and Indexing}
\label{sec:offline-pipeline}

The offline pipeline is responsible for transforming raw, fragmented engineering data into structured knowledge representations and indexing them for efficient semantic retrieval. It comprises four key stages: data acquisition and preprocessing, semantic annotation and structuring, embedding model adaptation, and vector indexing.

\subsubsection{Data Acquisition and Preprocessing} 
Raw data are acquired from internal engineering tools, including \textit{Software~1} (subsystem-level test sequences), \textit{Software~2} (component metadata and signal definitions), and \textit{Software~3} (requirement specifications). These tools expose data through APIs or structured exports when available. However, due to the lack of standardized programmatic access and inconsistent formats, a dedicated automation pipeline was developed to extract information. The pipeline incorporates standard text preprocessing (e.g., thresholding, layout normalization), followed by pattern-based parsing and heuristic field extraction. Each entry is enriched with metadata such as parent component, signal count, and timestamp, and stored in structured JSON format. 

To ensure robustness in industrial environments, the extraction pipeline supports logging, checkpointing, and recovery mechanisms. These features enable scalable, fault-tolerant operation, and consistent data collection despite interruptions or transient failures. The resulting corpus provides aligned requirement–test sequence pairs as the foundation for supervised training and retrieval.

\subsubsection{Semantic Structuring and Annotation} 
After preprocessing, the corpus is transformed into two structured formats:

\textbf{RAG Corpus (Knowledge Base):} 
    The knowledge base is a collection of JSON objects, each representing a semantically coherent retrieval unit.  
    Each object contains:
    \begin{itemize}
        \item \texttt{id}: a unique document identifier.
        \item \texttt{title}: a concise, functional descriptor.
        \item \texttt{requirements}: the main textual content.
        \item \texttt{description} (optional): supplementary metadata or clarifications.
        \item \texttt{sequences} (optional): associated test procedures.
        \item \texttt{category}: indicating the source or classification.
    \end{itemize}
    This corpus is indexed and stored in the vector database to support semantic retrieval.

\textbf{Triplet Dataset (Embedding Fine-Tuning)}: Created to facilitate embedding model adaptation via contrastive learning, each entry includes:
    \begin{itemize}
        \item \texttt{anchor}: requirement-like queries from real-world engineering tasks;
        \item \texttt{positive}: semantically related test sequences or documentation snippets;
        \item \texttt{negative}: domain-relevant but functionally unrelated items.
    \end{itemize}
    The triplets are generated through a hybrid approach: heuristic mining identifies candidate pairs based on shared signals, module IDs, and similarity; while LLM-assisted synthesis produces plausible positives and challenging (hard) negatives based on domain-specific prompts. A high-quality reference embedding model is also used to mine \textit{informative negatives} by ranking candidate sequences by intermediate similarity scores and selecting difficult but incorrect examples. This method improves generalization by encouraging the embedding model to distinguish subtle semantic differences between similar but functionally distinct artifacts.

\subsubsection{Embedding Model Fine-Tuning} 
General-purpose embedding models are not optimized for the specialized terminology and semantic relationships present in automotive testing. To address this, the embedding function is adapted to the domain using supervised fine-tuning with contrastive learning. A domain-specific triplet dataset is used to train the model to map relevant pairs closer and unrelated pairs farther apart in the embedding space. Both triplet loss and pairwise contrastive loss formulations are explored during training. The resulting fine-tuned model is deployed as a dedicated inference service to support downstream indexing and retrieval tasks within the RAG pipeline.

\subsubsection{Vector Indexing}
The domain-adapted embedding vectors generated from the corpus were stored and indexed in Azure Cosmos~DB for MongoDB, utilizing the native vector search extension (\texttt{\$vectorSearch}) for efficient approximate nearest neighbor retrieval. 

In this work, two indexing strategies were applied: \textit{dense vector indexing} for high-dimensional semantic retrieval, and \textit{metadata-based filtering indexing} to enable hybrid query capabilities based on attributes such as title and category. Additionally, the system supports incremental re-embedding to dynamically update and maintain retrieval accuracy as new engineering data becomes available.

The selection of Azure Cosmos~DB aligns with organizational mandates on data governance, infrastructure stability, scalability, and native integration with existing CI/CD pipelines. Alternative solutions, such as Qdrant, were considered but ultimately not pursued due to their higher operational overhead and reduced alignment with enterprise-scale requirements.

\subsection{Online Pipeline: Inference and Response Generation}
\label{sec:online-pipeline}

The online pipeline operationalizes the retrieval-augmented generation architecture, providing context-aware and traceable responses to engineering queries within automotive HIL testing workflows. This pipeline encompasses five detailed functional stages: (1) user query handling and preprocessing, (2) semantic retrieval of context documents, (3) dynamic context assembly and prompt generation, (4) hybrid reasoning and external tool integration, and (5) frontend integration and logging.

\subsubsection{User Query Handling and Preprocessing}
When engineers submit natural language queries via the frontend interface, queries undergo preprocessing to standardize input formats and minimize noise. The preprocessing steps include:
\textit{tokenization and normalization}, in which queries are tokenized using the SentencePiece tokenizer consistent with the fine-tuned embedding models, ensuring optimal embedding quality, and \textit{metadata extraction}, where keywords or specific domain terms (e.g., subsystem names, component identifiers) are extracted using heuristic rules or regular expressions, assisting in subsequent metadata-based filtering during retrieval.

\subsubsection{Semantic Retrieval of Context Documents}
The normalized and tokenized query is converted into a dense semantic embedding vector through the fine-tuned embedding service. This vector serves as input for approximate nearest-neighbor search against the indexed vector representations in the Azure Cosmos~DB vector store. Retrieval employs the built-in \texttt{\$vectorSearch} operator configured for cosine similarity search, returning the top-$k$ documents ranked by semantic relevance. 

The retrieval process first performs \textit{dense vector similarity} search to identify semantically relevant requirements or test sequences, followed by \textit{metadata-based filtering} (e.g., subsystem or component names) to refine results and ensure domain relevance.

\subsubsection{Dynamic Context Assembly and Prompt Generation}
Retrieved documents and associated metadata are assembled into a structured context buffer for input to the large language model. To address the LLM’s token limitation (e.g., GPT-4-turbo’s 8,192-token maximum), a priority-based truncation strategy sorts retrieved documents by semantic similarity and dynamically truncates or removes lower-ranked ones to retain the most relevant content within the prompt.

The context buffer is integrated into domain-specific prompt templates designed to guide the model's generation and maintain factual grounding. A representative prompt template is as follows:
\begin{quote}\small
\texttt{You are a specialized assistant supporting automotive HIL testing. Using only the following provided information, accurately answer the engineer's query:} \\
\texttt{---}\\
\texttt{[Retrieved Documents and Metadata]}\\
\texttt{---}\\
\texttt{Engineer’s Query: "[User Question]"}\\
\texttt{Response:}
\end{quote}

These templates are crafted through iterative refinement with domain experts to optimize the accuracy, relevance, and interpretability of model-generated responses.

\subsubsection{Hybrid Reasoning and External Tool Integration}
Beyond static retrieval, the HIL-GPT agent supports dynamic reasoning through structured external tool invocation. Leveraging Azure OpenAI’s function calling capability, the agent dynamically decides whether additional information from external tools is necessary to answer complex or context-specific queries accurately. In the agent implementation, this capability is realized through the integration of three categories of tools that augment the agent’s domain awareness and execution ability:
\begin{itemize}
    \item \textbf{Real-time test sequence retrieval}: Queries live databases (e.g., \textit{Software~1} or \textit{Software~3}) for current test case versions or recent updates.
    \item \textbf{CAN signal metadata queries}: Dynamically retrieves detailed signal descriptions and definitions from internal tools (e.g., \textit{Software~2}) to augment responses with accurate technical specifics.
    \item \textbf{Procedural action invocation}: Executes actions directly within the HIL environment, such as triggering visualizations or simulations on the test bench, based on natural language commands interpreted by the model.
\end{itemize}

When the LLM generates a structured tool invocation request, the backend system first parses and validates the requested action. It then routes the request through a standardized execution wrapper to ensure secure and consistent handling. Finally, the retrieved data or execution results are integrated back into the generation context to support precise, real-time problem-solving in complex engineering workflows.

\subsubsection{Frontend Integration and Logging}
The interaction with engineers is enabled through a React-based chatbot frontend connected to a FastAPI backend via WebSocket. The frontend supports interactive query submission with immediate, structured responses that include sources and traceable metadata, as well as real-time feedback capture, allowing users to rate responses, flag inaccuracies, and provide corrections or clarifications directly within the interface. 

All feedback is recorded in an Azure-managed database, storing the original user queries with their corresponding retrieved documents, along with the model-generated responses and related metadata such as retrieval scores and invoked tools. This feedback loop serves as a critical resource for system enhancement, supporting both qualitative and quantitative evaluation of model performance and enabling incremental retraining or embedding updates driven by actual user interactions.

\subsubsection{Deployment Modes} 
The system operates in two modes. In the \textit{offline engineering support mode}, engineers query requirements and test-related knowledge during development. In the \textit{online interactive control mode}, the agent is integrated with the HIL test bench to interpret engineer commands and translate them into display or test actions in real time. Together, these modes illustrate the system’s capability to connect human intent with machine-executable actions while ensuring full traceability.

\subsubsection{Traceability, Auditability, and Regulatory Compliance}
Ensuring full transparency and accountability, the system maintains a comprehensive audit trail logging each inference step, including retrieval results, model prompts, tool invocations, and final generated outputs. Logged entries contain timestamped references and unique identifiers linking user interactions, responses, and underlying data sources. This structured logging meets automotive industry regulatory requirements and provides traceability for auditing, debugging, and future system verification activities.

\section{Evaluation}
\label{sec:evaluation}

This section presents the experimental evaluation of the proposed system, structured around the following experimental research questions (ERQs):

\begin{itemize}
    \item \textit{\textbf{ERQ 1:}} To what extent does fine-tuning of embedding models improve retrieval performance in the HIL domain, and which embedding model performs best?
    \item \textit{\textbf{ERQ 2:}} How does triplet-based training with hard negative sampling affect the quality of embeddings?
    \item \textit{\textbf{ERQ 3:}} Does the integration of a RAG pipeline enhance response relevance and reduce hallucinations compared to retrieval-only and generation-only baselines?
    \item \textit{\textbf{ERQ 4:}} From the perspective of end-users, does HIL-GPT provide more relevant and trustworthy answers than a general-purpose LLM?    
\end{itemize}

\subsection{Evaluation Setup}
\label{sec:evaluation-setup}

The evaluation was designed to address the above ERQs through a combination of offline retrieval experiments, end-to-end agent assessment, and user-centered A/B testing. The benchmark dataset was constructed from internal requirements and test sequence records extracted from the automotive testing environment, aligned into semantically related pairs using heuristics based on component identifiers, signal references, and textual similarity. The dataset was further structured into triplets for contrastive evaluation, each consisting of an \textit{anchor} (requirement), a semantically relevant \textit{positive} (linked test sequence), and a \textit{negative} (unrelated or weakly related sequence).

Experiments were conducted in a unified inference environment with cosine similarity, FAISS-based vector indexing, and identical preprocessing logic. The evaluation process was organized into four experiments, each corresponding to one of the ERQs.

\paragraph{Exp 1: Impact of Fine-Tuning and Embedding Model Comparison.} 
To evaluate the benefit of fine-tuning and compare embedding backbones, several open-source embedding models from the SentenceTransformers~\cite{reimers-2019-sentence-bert} library were benchmarked. An initial screening was performed on a stratified 10\% subset (520 triplets) to eliminate underperforming candidates, followed by full-scale evaluation on the complete benchmark of 2,172 triplets. Retrieval performance was quantified using Top-1 accuracy.

\paragraph{Exp 2: Effectiveness of Triplet Training with Hard Negatives.}  
To assess whether triplet-based training with hard negative sampling improves embedding quality, two variants of the fine-tuned model were compared: one trained with randomly sampled negatives and one with hard negatives mined through a teacher model. Evaluation on the full benchmark measured the impact of hard negative mining on retrieval accuracy and robustness.

\paragraph{Exp 3: Benefits of RAG Integration.}  
To investigate whether RAG improves response quality and mitigates hallucinations, a set of 3,208 queries was constructed for which the Top-5 retrieved documents (using both original and fine-tuned embeddings) contained the correct answer. These were fed to \texttt{GPT-4o}~\cite{openai-gpt4o} and \texttt{GPT-4o-mini}~\cite{openai-gpt4o-mini} in two modes: with and without retrieved context. Document-level grounding accuracy was computed to quantify the benefit of RAG over generation-only and retrieval-only baselines.

\paragraph{Exp 4: User-Perceived Utility in A/B Testing.}  
To capture user perceptions, an A/B testing experiment was conducted with professional engineers. Participants were presented with anonymized answer pairs, one from HIL-GPT and one from a general-purpose LLM, for the same queries. They rated each answer in terms of relevance, trustworthiness, and traceability. Statistical analysis of the preferences provides insight into the end-user benefit of HIL-GPT in practical settings.

\subsection{Exp 1: Impact of Fine-Tuning and Embedding Model Comparison}
\label{sec:exp-embedding}

This experiment investigates the effect of domain-specific fine-tuning on embedding performance and compares open-source and proprietary embedding models under industrial constraints. The evaluation addresses \textit{ERQ~1}, assessing the retrieval accuracy, responsiveness to fine-tuning, and economic viability of candidate models in the context of automotive HIL testing. Given the relatively small user base typical of expert systems, the analysis prioritizes both performance and cost-efficiency.

\paragraph{Preliminary Evaluation on Benchmark Subset.}  
An initial screening was conducted on a stratified 10\% subset of the benchmark dataset (520 triplets) to identify promising open-source models. Nine models, spanning architectures and sizes, were compared using triplet accuracy. 

The triplet accuracy is used as the primary metric for evaluating the embedding models on the benchmark. Each sample in the benchmark consists of an anchor $a$, a semantically relevant positive $p$, and an irrelevant or weakly related negative $n$. The embedding model $f_\theta(\cdot)$ maps each text into a vector representation, and the cosine similarity between vectors is used to assess closeness. For a given triplet $(a, p, n)$, the prediction is considered correct if the similarity between the anchor and the positive is higher than that between the anchor and the negative:
\begin{equation}
\text{sim}(f_\theta(a), f_\theta(p)) > \text{sim}(f_\theta(a), f_\theta(n)).
\end{equation}

The triplet accuracy (TAcc.) is then computed as the proportion of correctly ranked triplets over the total number of triplets:
\begin{equation}
\begin{split}
\text{TAcc.} = \frac{1}{N} \sum_{i=1}^{N} 
\mathbb{1} \big[\, &\text{sim}(f_\theta(a_i), f_\theta(p_i)) \\
&> \text{sim}(f_\theta(a_i), f_\theta(n_i)) \,\big],
\end{split}
\end{equation}
where $N$ is the number of triplets and $\mathbb{1}[\cdot]$ denotes the indicator function.

As shown in \tablename~\ref{tab:triplet_eval_subset}, \texttt{gtr-t5-large}, \texttt{gtr-t5-xl}, and \texttt{bge-base-en-v1.5} exhibited the highest retrieval accuracy, and were selected for further evaluation.

\begin{table}[htbp]
\centering
\caption{Triplet accuracy (TAcc.) on 10\% benchmark subset.}
\label{tab:triplet_eval_subset}
\begin{tabularx}{\linewidth}{lrr}
\toprule
\textbf{Model} & \textbf{Params (M)} & \textbf{TAcc. (\%)} \\
\midrule
\texttt{gtr-t5-large}~\cite{gtr-t5-large}               & 335   & 84.29 \\
\texttt{gtr-t5-xl}~\cite{gtr-t5-xl}                     & 1241  & 83.01 \\
\texttt{bge-base-en-v1.5}~\cite{bge-base-en-v1.5}       & 110   & 80.91 \\
\texttt{multi-qa-mpnet-base-dot-v1}~\cite{multi-qa-mpnet-base-dot-v1} & 109 & 80.77 \\
\texttt{multi-qa-mpnet-base-cos-v1}~\cite{multi-qa-mpnet-base-cos-v1} & 109 & 79.49 \\
\texttt{all-mpnet-base-v2}~\cite{all-mpnet-base-v2}     & 109   & 78.85 \\
\texttt{all-distilroberta-v1}~\cite{all-distilroberta-v1} & 82   & 77.24 \\
\texttt{all-mpnet-base-v1}~\cite{all-mpnet-base-v1}     & 109   & 77.56 \\
\texttt{all-roberta-large-v1}~\cite{all-roberta-large-v1} & 355  & 75.64 \\
\bottomrule
\end{tabularx}
\end{table}

\paragraph{Full-Scale Zero-Shot Evaluation.}  
The top three open-source models identified in the initial screening were further evaluated on the full benchmark of 2,172 queries. For reference, two proprietary models (\texttt{text-embedding-ada-002}~\cite{text-embedding-ada-002} and \texttt{text-embedding-3-small}~\cite{text-embedding-3-small}) from OpenAI were also included in the comparison. The parameter counts of the proprietary OpenAI models, \texttt{text-embedding-ada-002} and \texttt{text-embedding-3-small}, have not been publicly disclosed. Community estimates place \texttt{text-embedding-ada-002} in the range of several hundred million parameters and \texttt{text-embedding-3-small} in the range of roughly 1–3 billion parameters.

In this experiment, the Top-1 retrieval accuracy was used. It measures the proportion of queries for which the system correctly ranks the relevant target document at the first position in the retrieved list. Let the benchmark dataset consist of \(N\) queries, each associated with a ground truth target document \(d_i^\text{true}\). For each query \(q_i\), the system returns an ordered list of candidates \(\{d_i^{(1)}, d_i^{(2)}, \dots\}\), where \(d_i^{(1)}\) denotes the top-ranked document. Define an indicator function:
\begin{equation}
\mathbb{1}_{i} =
\begin{cases}
1, & \text{if } d_i^{(1)} = d_i^\text{true} \\
0, & \text{otherwise}
\end{cases}
\end{equation}

Then the Top-1 retrieval accuracy is computed as:
\begin{equation}
\text{Top-1 Accuracy} =
\frac{1}{N} \sum_{i=1}^{N} \mathbb{1}_{i}
\end{equation}

This metric provides a strict evaluation of the retrieval system by only considering a query as correct if the most relevant document is ranked first. Compared to more relaxed metrics such as Top-$k$ accuracy (\(k>1\)), it offers a clear and interpretable measure of first-choice correctness.

As shown in \tablename~\ref{tab:embedding_model_full}, the proprietary models achieved higher top-1 accuracy (\texttt{text-embedding-ada-002}: 58.89\%, \texttt{text-embedding-3-small}: 58.70\%) than the best-performing open-source model (\texttt{gtr-t5-large}:~58.24\%). Notably, the largest open-source model (\texttt{gtr-t5-xl}, 1.24B parameters) did not surpass smaller variants, suggesting that increased model size alone does not guarantee improved retrieval in this domain. The compact \texttt{bge-base-en-v1.5} demonstrated competitive performance (50.69\%), making it an appealing candidate for fine-tuning due to its favorable efficiency–accuracy trade-off.

\begin{table}
\centering
\caption{Top-1 retrieval accuracy on full benchmark (2,172 queries).}
\label{tab:embedding_model_full}
\begin{tabular}{@{}lrr@{}}
\toprule
\textbf{Model} & \textbf{Params (M)} & \textbf{Top-1 Acc. (\%)} \\
\midrule
\texttt{text-embedding-ada-002} & --- & 58.89 \\
\texttt{text-embedding-3-small} & --- & 58.70 \\
\texttt{gtr-t5-large} & 334.94 & 58.24 \\
\texttt{gtr-t5-xl} & 1240.91 & 57.32 \\
\texttt{bge-base-en-v1.5} & 110.00 & 50.69 \\
\bottomrule
\end{tabular}
\end{table}

\begin{figure}[htbp]
    \centering
    \includegraphics[width=0.95\linewidth]{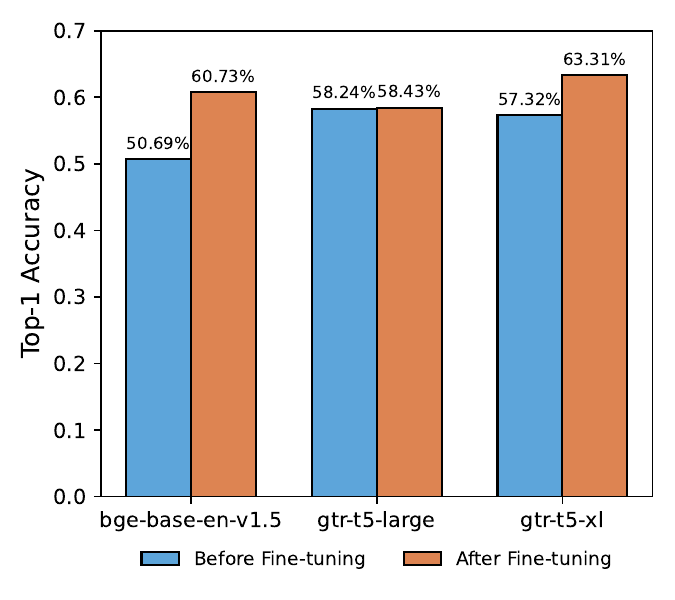}
    \caption{Pre- and post-fine-tuning accuracy comparison (Top-1 retrieval).}
    \label{fig:fine_tuning_comparison}
\end{figure}

\paragraph{Effect of Fine-Tuning.}  
As of this work, proprietary embedding models such as \texttt{text-embedding-ada-002} cannot be fine-tuned by end-users. The models are only accessible as pre-trained inference APIs without support for domain-specific adaptation. Thus, to quantify the impact of domain-specific adaptation, the three open-source models were fine-tuned using the curated triplet dataset with contrastive loss. 

\figurename~\ref{fig:fine_tuning_comparison} summarizes the results. Fine-tuning significantly improved \texttt{bge-base-en-v1.5} by over 10 percentage points (from 50.69\% to 60.73\%), while \texttt{gtr-t5-xl} achieved the highest absolute accuracy (63.31\%) with moderate improvement. The mid-sized \texttt{gtr-t5-large} showed negligible gains.

\begin{table}
\centering
\caption{Estimated fine-tuning cost and inference latency of three models.}
\label{tab:ft_latency}
\begin{tabular}{@{}lrr@{}}
\toprule
\textbf{Model} & \textbf{Fine-Tuning Cost (USD)} & \textbf{Latency (ms)} \\
\midrule
\texttt{bge-base-en-v1.5} & $\sim$50  & $\sim$15 \\
\texttt{gtr-t5-large}     & $\sim$150 & $\sim$45 \\
\texttt{gtr-t5-xl}        & $\sim$600 & $\sim$120 \\
\bottomrule
\end{tabular}
\end{table}

\tablename~\ref{tab:ft_latency} summarizes the estimated fine-tuning cost and inference latency of the three evaluated models after adaptation. The results indicate that while the large \texttt{gtr-t5-xl} achieves the highest absolute accuracy, its substantial fine-tuning cost and inference latency limit its practicality in industrial settings. Conversely, the compact \texttt{bge-base-en-v1.5} benefits most from fine-tuning, likely due to underfitting before domain adaptation, and achieves a favorable balance of accuracy, cost, and efficiency. This aligns with capacity–data matching theory, which suggests diminishing returns when fine-tuning large models on limited domain data. Given these considerations, \texttt{bge-base-en-v1.5} was selected as the final embedding model for deployment.

In response to \textbf{\textit{ERQ~1}}, the results demonstrate that fine-tuning improves the adaptation of general-purpose embeddings to specialized engineering domains. In few-shot settings, smaller models offer a better trade-off between performance, efficiency, and cost, emphasizing the importance of aligning model capacity with data availability and industrial constraints. Based on these considerations and the observed improvements after fine-tuning, \texttt{bge-base-en-v1.5} was selected as the embedding backbone for the final system.

\subsection{Exp 2: Effectiveness of Triplet Training with Hard Negatives}
\label{sec:negative-samples}

Contrastive learning typically relies on triplets consisting of an anchor, a positive, and a negative sample. However, in industrial environments, particularly in safety-critical domains such as automotive testing, curated negative examples are often unavailable, as engineers usually maintain only positive associations between requirements and corresponding test sequences without formally annotating unrelated pairs.

To investigate the impact of this constraint, an experiment was conducted on the \texttt{bge-base-en-v1.5} model. Two fine-tuning variants were trained: one using the full triplet dataset (including negative samples), and another using only anchor–positive pairs, effectively removing explicit contrastive supervision. 

\begin{figure}[htbp]
    \centering
    \includegraphics[width=1\linewidth]{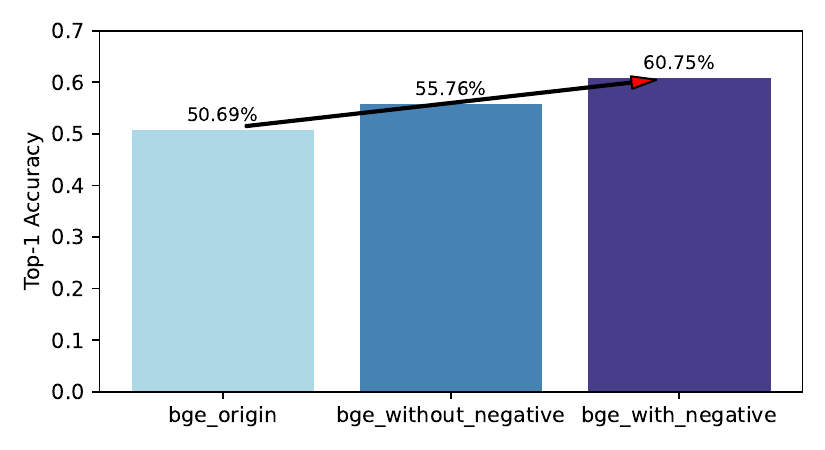}
    \caption{Top-1 retrieval accuracy of \texttt{bge-base-en-v1.5} under different training regimes.}
    \label{fig:bge_negative_ablation}
\end{figure}

\figurename~\ref{fig:bge_negative_ablation} reports the top-1 retrieval accuracy across three regimes: pre-trained (no fine-tuning), fine-tuned without negatives, and fine-tuned with full triplets. The model fine-tuned without negatives reached 55.76\% accuracy, improving upon the baseline 50.69\%, but still below the 60.75\% achieved with full triplet-based fine-tuning.

These findings address \textbf{\textit{ERQ~2}}, demonstrating that triplet-based training with hard negative sampling improves the quality of learned embeddings by enhancing their ability to separate semantically similar from unrelated content. Even under limited supervision, domain-specific fine-tuning without negatives yields moderate improvements, indicating that embedding models can adapt to specialized tasks with only positive examples. However, the absence of negatives weakens the model’s discriminative power, as reflected by lower retrieval accuracy. The results, therefore, confirm the importance of contrastive signals provided by negative samples. In industrial environments where negative examples are scarce or costly to produce, incorporating a small set of synthetic negatives into fine-tuning can substantially improve embedding quality. This supports the adoption of hybrid fine-tuning strategies that combine weak supervision with curated contrastive signals, enabling practical and effective adaptation of embedding models within enterprise constraints.

\subsection{Exp 3: Impact of RAG on Response Quality and Hallucination Mitigation}

This subsection addresses \textbf{\textit{ERQ~3}}: Does the integration of a RAG pipeline enhance response relevance and reduce hallucinations compared to retrieval-only and generation-only baselines?

The experiment evaluated how improvements in embedding quality translate into downstream LLM behavior when operating in a RAG setting. Specifically, the experiment examined whether fine-tuned embeddings enable the language model to more accurately ground its responses in the retrieved documents and reduce hallucinated content.

Two LLMs, \texttt{GPT-4o} and \texttt{GPT-4o-mini}, were tested on 1,700 domain-specific queries. For each query, the Top-5 documents were retrieved using either the original \texttt{bge-base-en-v1.5} embeddings or the fine-tuned \texttt{bge-base-en-v1.5-with-negative} embeddings. The LLMs were instructed to answer the query and explicitly return the ID of the source document used to derive the response. Accuracy was defined as the proportion of queries for which the LLM correctly attributed its answer to the relevant document in the provided context.

\begin{figure}[htbp]
    \centering
    \includegraphics[width=1\linewidth]{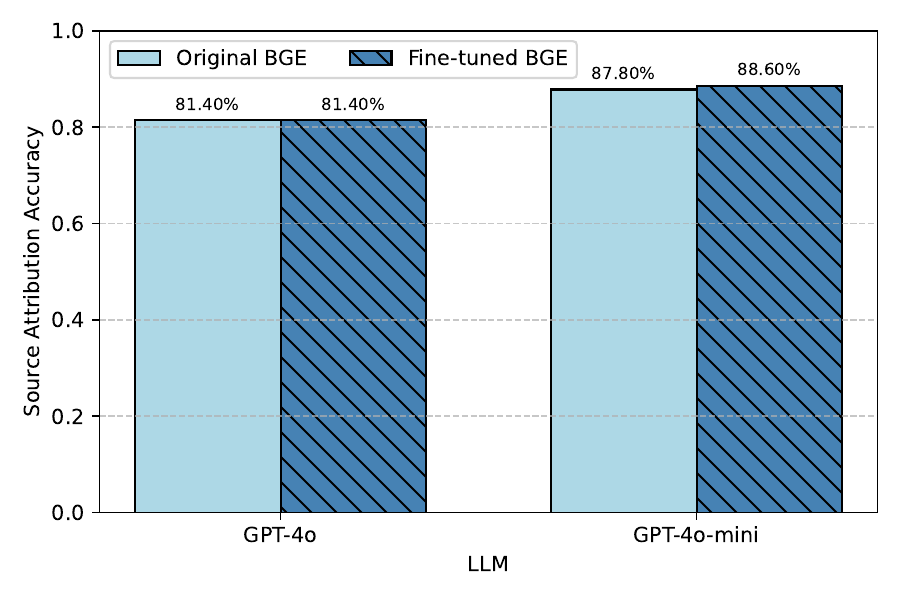}
    \caption{Comparison of source attribution accuracy between \texttt{GPT-4o} and \texttt{GPT-4o-mini} using original and fine-tuned \texttt{bge-base-en-v1.5} embeddings.}
    \label{fig:llm_vs_embedding_eval}
\end{figure}

As summarized in \figurename~\ref{fig:llm_vs_embedding_eval}, both LLMs achieved higher attribution accuracy when paired with fine-tuned embeddings.  \texttt{GPT-4o-mini} outperformed \texttt{GPT-4o} in grounding its answers to specific documents.With fine-tuned embeddings, \texttt{GPT-4o-mini} improved to 88.6\%, whereas \texttt{GPT-4o} showed minimal improvement.

These findings demonstrate that the integration of a RAG pipeline with domain-adapted embeddings enhances response relevance and mitigates hallucinations compared to generation-only approaches. Furthermore, the results suggest that larger LLMs do not necessarily yield better grounding behavior in this domain. The higher capacity of \texttt{GPT-4o} may lead to overgeneralization or context diffusion, whereas the more compact \texttt{GPT-4o-mini} maintains a sharper focus and better adherence to prompt constraints.  

From a design perspective, optimizing the embedding model for precise domain retrieval reduces the cognitive burden on the LLM and improves the controllability of generated outputs. Additionally, selecting smaller LLMs for post-retrieval reasoning offers advantages in latency, cost, and interpretability without compromising grounding fidelity.

These observations confirm that RAG pipelines, particularly when combined with fine-tuned embeddings, enhance the factual consistency and traceability of responses in industrial testing environments, directly addressing the concerns raised in \textbf{\textit{ERQ~3}}.

\subsection{User-Perceived Utility of RAG-Augmented Agent}
\label{sec:erq4}

To address \textbf{\textit{ERQ 4}}, whether end-users perceive a RAG-augmented agent as more relevant and trustworthy than a general-purpose LLM, this study conducted an A/B user evaluation in a realistic engineering context.

Ten engineers from two functional domains (Operation and Exterior Function) interacted with two chatbot variants: Bot A, equipped with RAG, and Bot B, a traditional generative chatbot without retrieval support. Each participant completed identical tasks on both interfaces, without knowledge of the underlying configurations. After each interaction, participants provided feedback on several dimensions:
\begin{itemize}
    \item Helpfulness (forced choice between Bot A and Bot B)
    \item Ratings of completeness, truthfulness, naturalness, and overall satisfaction (1–5 scale)
    \item Open-ended comments and suggestions
\end{itemize}

Additionally, free-text feedback was analyzed for sentiment polarity (range: -1 to +1) using a transformer-based sentiment classifier~\cite{wolf2020transformers}.

\begin{figure}[htbp]
    \centering
    \includegraphics[width=1\linewidth]{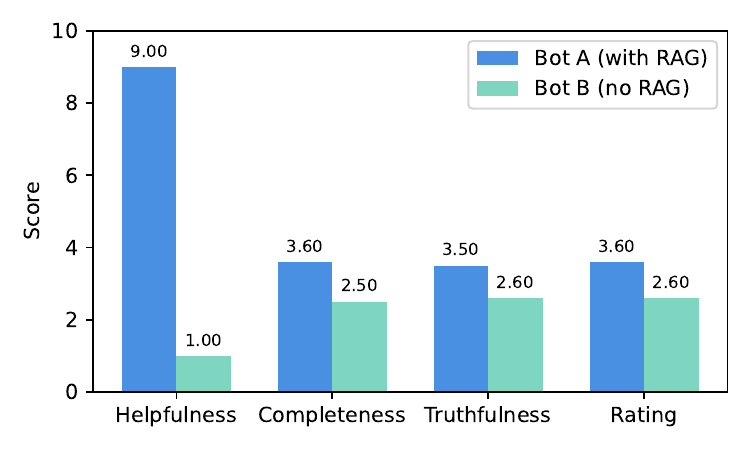}
    \caption{User evaluation ratings: Bot A (with RAG) vs. Bot B (no RAG).}
    \label{fig:ab_rating_stats}
\end{figure}

Bot A (with RAG) was preferred as more helpful in 9 out of 10 sessions. Figure~\ref{fig:ab_rating_stats} shows average user ratings: Bot A achieved higher scores in completeness (3.6 vs 2.5), truthfulness (3.5 vs 2.6), and overall satisfaction (3.6 vs 2.6).

Participants described Bot A as “more aligned with operations context,” “explained automation logic clearly,” and “handled signal-level details with actionable advice.” Some acknowledged Bot B’s faster response time, but noted that its answers were less complete and less relevant to domain needs. In one dissenting case, a participant favored Bot B for its brevity and quicker responses (2.4 s for Bot B vs 17.4 s for Bot A on average), emphasizing that testers sometimes prefer concise, directly actionable output over richer explanations.

The qualitative analysis indicates that RAG-augmented responses enhance perceived relevance and trust, although at the expense of higher latency. This trade-off may be acceptable in engineering contexts where traceability and completeness are valued over minimal response time.

These findings support the hypothesis of \textbf{\textit{ERQ~4}} that RAG-augmented agents deliver higher user-perceived utility in domain-specific environments, confirming the relevance and trustworthiness of grounded answers compared to general-purpose generation. The results also highlight practical trade-offs between response latency and output richness, suggesting opportunities to tailor response style based on task type and user preferences.

\section{Lessons Learned and Practical Insights}
\label{sec:dis}
This section reflects on key findings from the evaluation, discussing performance patterns, system limitations, and implications for industrial deployment.

\textbf{A. Embedding Model Performance} The experiments showed that model size alone is not a reliable predictor of retrieval performance. Smaller models such as \texttt{gtr-t5-large} outperformed the larger \texttt{gtr-t5-xl} in the retrieval-based setting, while the compact \texttt{bge-base-en-v1.5} achieved competitive results once fine-tuned. This underscores the importance of domain alignment and fine-tuning over raw capacity. Fine-tuning improved \texttt{bge-base-en-v1.5} by over 10 percentage points in top-1 accuracy, confirming that smaller, task-optimized models adapt more effectively in constrained domains.

\textbf{B. Retrieval and Generation Trade-offs} Improved embeddings also enhanced downstream LLM response grounding. Fine-tuning the embedding model increased correct attribution rates in both GPT variants, with \texttt{GPT-4o-mini} outperforming the larger \texttt{GPT-4o}. This suggests that high-capacity LLMs may generalize too broadly, while smaller models maintain focus, which is beneficial when traceability is critical.

\textbf{C. Negative Sampling in Fine-Tuning} The ablation study confirmed that negative samples are essential for effective contrastive learning. Removing negatives reduced top-1 retrieval accuracy by 5 points, illustrating their role in sharpening semantic boundaries. Hard negatives selected by mid-range similarity proved particularly effective, supporting the use of curated or synthetic negatives in low-data settings.

\textbf{D. Deployment Considerations} Fine-tuned \texttt{bge-base-en-v1.5} offered the best trade-off between accuracy, cost, and latency, making it well-suited for real-time industrial use. Larger models provided marginal gains at significantly higher resource costs. Similarly, smaller LLMs achieved better attribution consistency than their larger counterparts, highlighting the need to align model choice with task constraints.

\textbf{E. Generalization and Usability} The fine-tuned system generalized well across functional domains, though performance degraded in loosely structured, user-facing modules with informal language. User feedback favored the RAG-enabled agent for its relevance, accuracy, and use of familiar terminology, but also highlighted the need for concise, actionable outputs.

In summary, effective engineering assistants require co-optimization of embedding and generative components, balanced against deployment constraints and user expectations.

\section{Summary and Future Work}
\label{sec:summary}
This paper presented \textbf{HIL-GPT}, a retrieval-augmented intelligent assistant tailored for automotive HIL testing environments. Through a systematic evaluation of open- and closed-source embedding models, fine-tuning strategies, and retrieval-augmented generation (RAG), the findings demonstrate that larger models do not necessarily yield superior performance for domain-specific, retrieval-centric tasks. Instead, compact models, when properly fine-tuned and integrated with RAG, achieve a favorable trade-off between retrieval accuracy, latency, and deployment cost, making them more suitable for industrial use cases.

For future work, several extensions are envisioned. Large-scale user evaluations across more diverse HIL domains can validate long-term usability and generalization. Exploring hybrid few-shot strategies and adaptive negative sampling may further improve embedding robustness in low-resource scenarios. Finally, integrating additional structured signals and graph-based reasoning could enhance the system’s ability to handle loosely defined or user-centric requirements, further advancing automated test case generation in safety-critical engineering settings.

\section*{Declaration of Competing Interest}
The authors declare that they have no known competing financial interests or personal relationships that could have appeared to influence the work reported in this paper. 

\section*{CRediT authorship contribution statement}
\textbf{Chao Feng.} Methodology, Conceptualization, Writing, Review \& Editing.
\textbf{Zihan Liu.} Methodology, Writing, Data curation, Review. 
\textbf{Siddhant Gupta.} Methodology, Review. 
\textbf{Gongpei Cui.} Methodology, Review. 
\textbf{Jan von der Assen.} Writing, Review \& Editing.
\textbf{Burkhard Stiller.} Supervision, Funding acquisition.

\section*{Acknowledgements}
This work has been supported by the University of Zürich UZH and by Volvo Cars, which funded and supported the project. Volvo Cars provided AI training data, AI training infrastructure, and expertise in automotive HIL testing.

\section*{Disclaimer}
The views and opinions expressed are those of the authors and do not necessarily reflect the official policy or position of Volvo Cars.

\bibliographystyle{cas-model2-names}
\bibliography{main}
\balance
\end{document}